\documentclass[10pt,twocolumn,letterpaper]{article}

\usepackage{iccv}
\usepackage{times}
\usepackage{epsfig}
\usepackage{graphicx}
\usepackage{amsmath}
\usepackage{amssymb}

\usepackage{graphicx}
\usepackage{amsmath}
\usepackage{amssymb}
\usepackage{booktabs}
\usepackage{multirow}
\usepackage{wrapfig}
\usepackage[table,x11names]{xcolor}

\definecolor{pearDark}{HTML}{2980B9}
\usepackage[pagebackref=true,breaklinks=true,letterpaper=true,colorlinks,bookmarks=false]{hyperref}

\iccvfinalcopy 

\usepackage[capitalize]{cleveref}
\crefname{section}{Sec.}{Secs.}
\Crefname{section}{Section}{Sections}
\Crefname{table}{Table}{Tables}
\crefname{table}{Tab.}{Tabs.}
\crefname{equation}{Eq.}{Eqs.}

\def\modelName{ContraWarping}

\begin{document}

\title{Unsupervised Facial Expression Representation Learning \\ with Contrastive Local Warping}

\author{First Author\\
Institution1\\
Institution1 address\\
{\tt\small firstauthor@i1.org}
\and
Second Author\\
Institution2\\
First line of institution2 address\\
{\tt\small secondauthor@i2.org}
}

\author{
Fanglei Xue$^{1}$\thanks{Work was down when Fanglei Xue was an intern at Baidu Research.} \; 
Yifan Sun$^{2}$ \;
Yi Yang$^{3}$\thanks{Corresponding author.}\\
$^{1}$ University of Technology Sydney \hspace{1mm} $^{2}$ Baidu Inc. \hspace{1mm} $^{3}$ Zhejiang University \\
{\tt\small xuefanglei19@mails.ucas.ac.cn, sunyf15@tsinghua.org.cn, yangyics@zju.edu.cn}
}

\maketitle

\begin{abstract}
This paper investigates unsupervised representation learning for facial expression analysis. We think Unsupervised Facial Expression Representation (UFER) deserves exploration and has the potential to benefit facial expression analysis regarding some critical problems, \eg scaling, annotation bias, the gap between discrete annotations and continuous emotion expressions, and model pre-training. Such motivated, we propose a UFER method with contrastive local warping (\modelName{}), which leverages the insight that the emotional expression is robust to current global transformation (affine transformation, color jitter, etc.) but can be easily changed by random local warping. Therefore, given a facial image, \modelName{} employs some global transformations and local warping to generate its positive and negative samples and sets up a novel contrastive learning framework. Our in-depth investigation shows that: 1) the positive pairs from global transformations may be exploited with general self-supervised learning (\eg BYOL) and already bring some informative features, and 2) the negative pairs from local warping explicitly introduce expression-related variation and further bring substantial improvement. Based on \modelName{}, we demonstrate the benefit of UFER under two facial expression analysis scenarios: facial expression recognition and image retrieval. For example, directly using \modelName{} features for linear probing achieves 79.95\% accuracy on RAF-DB, significantly reducing the gap towards the full-supervised counterpart (89.18\% / 84.81\% with/without pre-training).
\end{abstract}

\section{Introduction}
Facial expression is one of the most natural ways for humans to express their emotions by moving their facial muscles~\cite{darwin1998expression}. Facial Expression Analysis (FEA) aims at automatically analyzing the emotion from facial images and has wide applications in various domains, such as driver fatigue monitoring, virtual reality, human-computer interaction systems, \etc. In the last decades, FEA has made great progress benefiting from deep learning methods~\cite{cai2018IslandLoss,zeng2018facial,wang2020region,wang2020suppressing,li2021adaptively,xue2021TransFER}.
However, almost all of these methods rely on supervised learning, which requires large-scale and high-quality labeled datasets. Such datasets are scarce and expensive to obtain for FEA, which limits the performance of deep learning methods that can benefit from scaling up the training data. In addition, different datasets may have considerable annotation bias, leading to supervision conflict for joint training~\cite{zeng2018facial}. 
Therefore, in this paper, we are interested in unsupervised representation learning for facial expression analysis.

\begin{figure}[t]
  \centering
   \includegraphics[width=0.96\linewidth]{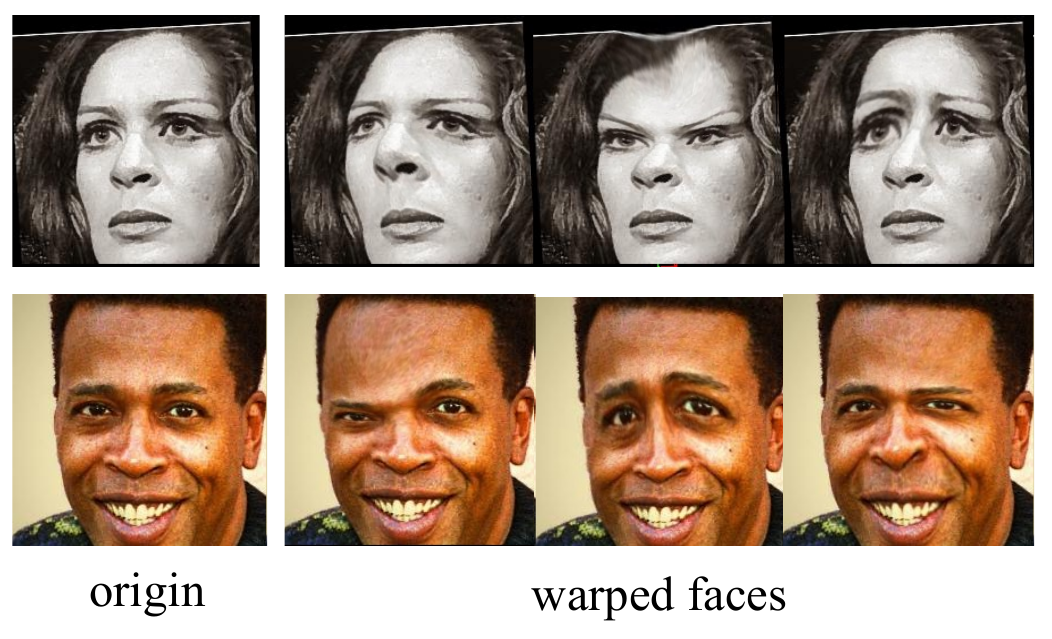}
   \caption{The motivation of our ~\modelName{}.
   We propose a random local warping to mimic facial muscle movements. 1st row: some movements may change the original expression (neutral) to some other predefined categories (sad, angry, and pitiable). 2nd row: some movements result in unreal expression, but they do change the expression (\eg action units) to some extent, as well.
   Since random local warping does not rely on human annotation, we utilize it for unsupervised facial expression representation learning to learn expression-related features. }
   \label{fig:intro}
\end{figure}

Besides the advantage of strong scaling capability, we think Unsupervised Facial Expression Representation (UFER) is potential to benefit automatic facial expression analysis in more aspects, such as the gap between discrete annotations and continuous emotion expressions, and model pre-training. We explain these two aspects as below:

$\bullet$ The gap between discrete annotations and continuous emotion expressions. One of the most popular FEA tasks is facial expression recognition (FER). It typically categorizes facial emotion into several (\eg 7) classes. Such discrete categorization is not consistent with the continuous variation of facial expressions. In some realistic facial expression analysis tasks (\eg expression retrieval and photo album summarization), the continuous feature space is superior than a discrete one \cite{vemulapalli2019CompactEmbedding}. UFER does not need human annotations and naturally bridges the gap between discrete annotations and continuous emotion expressions.

$\bullet$ Model pre-training is critical for facial expression analysis. For example, most facial expression recognition methods based on deep learning use pre-trained weights on MS1M~\cite{guo2016ms1m} or ImageNet~\cite{deng2009ImageNet} for model initialization. Without these pre-trained weights, the performance will drop significantly.
However, pre-training on ImageNet classification or face recognition deviates far from the objective of facial expression analysis. For example, in order to identify a same person under different scenes, a MS1M pre-trained model should focus on identity-related features and suppress the expression-related features. In contrast, we believe using UFER for model pre-training is likely to achieve better effect.

Such motivated, we propose an UFER method (\modelName{}) with contrastive local warping, inspired by recent generic self-supervised learning (SSL) methods (\eg MoCo \cite{He2020MoCo}, BYOL \cite{grill2020BYOL}, SimSiam \cite{Chen2020SimSIam}) based on contrastive learning. 
Generally, these contrastive SSL methods generate positive pairs from different views of a same image, and the negative pairs from different images (some SSL methods do not have negative pairs). Using contrastive learning, they train a deep feature space where the positive samples are close to each other and the negative samples are far away. 

Based on these general contrastive SSL methods, the key insight of our \modelName{} is: the emotion expression is robust to current widely-used data augmentations (marked as global transformations) like affine transformation, color jitter, \etc, but can be easily changed by random local warping. Therefore, given a facial image, \modelName{} employs some global transformations and local warping to generate its positive and negative samples, respectively. Based on these triplet samples, we set up a novel contrastive learning framework. 
Specifically, given a face image, we randomly select a region and generate local warping by moving the content to a random direction with a random distance. We find such random local warping 1) sometimes can simulate the realistic facial muscle movements and roughly change the facial expression to another realistic one (the first row in Fig.~\ref{fig:intro}), and 2) sometimes changes the facial expression to some unreal (and ridiculous) expression (the second row in Fig.~\ref{fig:intro}). No matter which situation happens, the locally-warped face is likely to have a different expression and thus becomes a negative sample for the original image. Given these negative samples, \modelName{} pushes them far away from the original image while pulling the positive samples close. 

To further enhance \modelName{}, we incorporate a facial landmark detection sub-task. 
We use an existing landmark detection method to extract pseudo-landmarks for the given face and perform the same warping operation to generate landmarks for the warped face. Since our goal is not to predict precise landmarks but to help the model to find moving muscles, no refined human-labelled landmark is needed here. Thus, our proposed framework could extract expression-related features in the pre-training stage without any annotations.

Based on \modelName{}, we conduct in-depth investigations on UFER and reveal that: 1) the positive pairs from global transformations may be exploited with general self-supervised learning (\eg BYOL) and already bring some informative features; and 2) the negative pairs from local warping explicitly introduce expression-related variation and further bring substantial improvement. Experiments on facial expression recognition and retrieval tasks validate the effectiveness of \modelName{}.

To summarize, our contributions are as follows:

\begin{enumerate}
  \item We propose a novel framework, \modelName{}, for unsupervised facial expression representation learning. It leverages random local warping to simulate facial muscle movements and generate informative negative pairs for contrastive learning.
  \item We introduce a facial landmark detection sub-task based on pseudo labels to help the model identify the expression-changing muscles, significantly improving the k-NN performance.

  \item Based on \modelName{}, we comprehensively investigate UFER against the supervised counterpart and reveal its strong potential for facial expression analysis.  
\end{enumerate}

Codes and pre-trained weights will be public at \url{https://github.com/youqingxiaozhua/ContraWarping}.

\begin{figure*}
  \centering
   \includegraphics[width=0.98\linewidth]{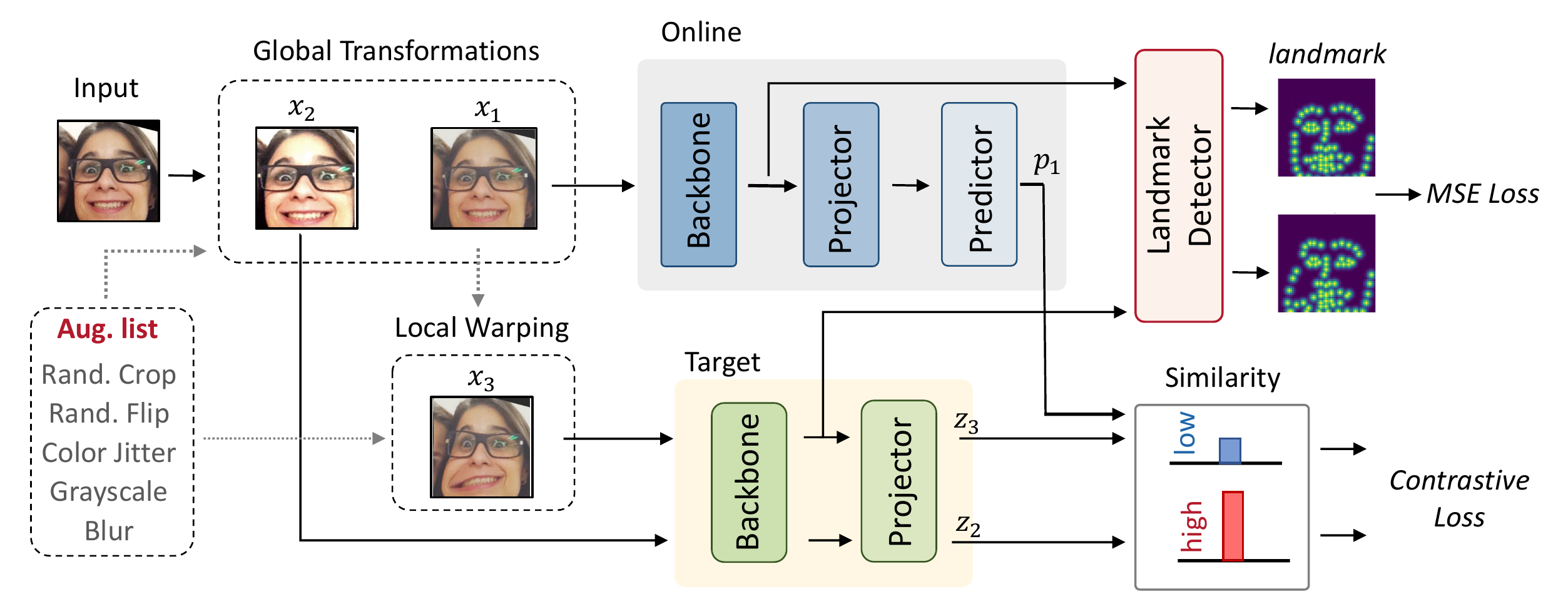}

   \caption{Overview of our \modelName{} framework. The model takes an input image and applies global transformations to produce $x_1$ and $x_2$, which should be similar in feature space. Then, it applies random local warping to $x_1$ to generate $x_3$, which simulates muscle movements. The model learns to push $x_3$ away from $x_1$ and $x_2$ in feature space, thus capturing expression-related features without human annotation. Additionally, a landmark detection sub-task is introduced to help the model identify the warped areas.}
   \label{fig:model}
\end{figure*}

\section{Related Works}
\subsection{Facial Expression Recognition}
FER methods tend to extract informative and expression-related features from facial images, and then adopt a classifier (\eg SVM~\cite{cortes1995support}) to classify the image into expression categories. In the last decades, many hand-crafted filters are proposed to extract texture-based features, like: LBP \cite{shan2009facial}, Gabor~\cite{liu2002gabor}, HOG \cite{dalal2005histograms}, and SIFT \cite{ng2003sift}. These methods could handle in-the-lab databases, but fail to extend to in-the-wild scenes due to various poses and occlusions.

Recently, benefiting from large-scale in-the-wild databases~\cite{li2017reliable,mollahosseini2017affectnet,barsoum2016training}, learning-based methods have made a great process for FER. Deep convolutional neural networks (CNN) have a strong ability to automatically extract discriminative features from images with supervision from ground-truth labels. Li \etal~\cite{li2017reliable} proposed the DLP-CNN to further enhance the discriminative power of deep features with a locality preserving loss. Cai~\etal~\cite{cai2018IslandLoss} also proposed an island loss to reduce the intra-class various and enlarge the inter-class differences. Ruan~\etal~\cite{ruan2021feature} proposed the FDRL method to first decompose facial features into action-aware latent features and then reconstruct the expression-specific features.

Those methods extract holistic features from the whole face, while other methods try to find expression-related facial areas to enhance recognition. Zhong~\etal~\cite{zhong2012LearningActiveFacial} proposed a two-stage multi-task sparse learning framework to find common patches shared by all expressions and specific patches to discriminate a certain expression. For the first time, they proved that only a few facial muscles (areas) are discriminative for FER. Haapy~\etal~\cite{happy2014automatic} proposed a method to extract some salient patches containing discriminative features with the help of facial landmarks. FER in the wild need to handle unconstrained conditions like partial occlusion and various poses. Li~\etal~\cite{Li2019OcclusionAware} proposed a gate-based method named ACNN, which utilizes the attention mechanism to compute an adaptive weight for every facial region. With the help of the proposed gate unit, ACNN could shift the attention from occluded patches to other unoccluded ones. 
RAN~\cite{wang2020region} utilizes self-attention and relation-attention modules to extract compact face representation from several face regions. Most recently, TransFER~\cite{xue2021TransFER} is proposed to utilize Transformer~\cite{vaswani2017attention} and multiple attention maps to learn relation-aware local features.

Collecting large-scale FER datasets' annotations is challenging due to high ambiguity and subjectivity. Therefore, another line of FER research is learning with uncertain or noisy labels. Zeng~\etal~\cite{zeng2018facial} firstly proposed the IPA2LT framework to learn from inconsistently labelled FER datasets. After that, many methods~\cite{wang2020suppressing,sheDiveAmbiguityLatent2021,zhang2021RelativeUncertaintya,zhang2022LearnAll} are proposed to decrease uncertainty.
Wu~\etal~\cite{wu2021NGCUnified} studied a new problem, learning with open-world noisy data and proposed a graph-based method to solve this problem.

Unlike these methods, we aim to learn general expression-aware features without any clean or noisy labels. We randomly generate various facial images with simulated muscle movements and push the model to focus on these regions to extract expression-related features.

\subsection{Learning from Unlabeled Data}
To reduce the dependency on high-cost annotated datasets, a large number of methods have been proposed to learn from unlabeled data. Among them, self-supervised learning methods have achieved great success in the last decade. He~\etal~\cite{He2020MoCo} proposed the MoCo framework with a momentum encoder which firstly outperforms supervised pre-training in some downstream tasks, indicating the great potential of contrastive learning.
SimCLR~\cite{chen2020SimpleFramework} and MoCo V2~\cite{chenMoCoV22020} further simplify this framework by removing the memory bank and adding a projection head after the representations.
Without negative samples or momentum encoder, BYOL~\cite{grill2020BYOL} and SimSiam~\cite{Chen2020SimSIam} make the framework more simpler and cleaner. 

Some methods use similar ideas to extract general facial representation from unlabeled data. SSSPL~\cite{shu2021LearningSpatialSemantic} adopts three auxiliary tasks (patch rotation, segmentation and classification tasks) to learn the spatial-semantic relationship. He~\etal~\cite{he2022EnhancingFace} utilized a 3D reconstruction task as a self-supervised bypass to enhance face recognition. TCAE~\cite{li2019TCAE} and FaceCycle~\cite{zhang2021LearningFacial} utilize multiple encoder and decoder to disentangle and reconstruct pose, expression, or identity features to learn from unlabeled data. TCAE also requires video samples to provide variations in expression and pose~\cite{li2019TCAE}.
As for the FER task, very limited related research focus on this topic. 
CRS-CONT~\cite{li2022CRSCONTWellTrainedb} adopt the self-supervised learning framework to FER. However, it still need coarse-grained labels to generate expression-specific positive and negative sample pairs. Differently, we proposed a random warping strategy to simulate the emotion expression process -- muscle movements, which could easily generate various expression-specific negative samples without supererogatory encoder-decoder. With our proposed \modelName{}, many existing self-supervised methods could be utilized to extract expression features without any labelled data.

\section{Method}

\subsection{Overview}
As has been discussed before, contrastive learning methods could learn without labels by producing highly similar representations for different views of the same image. Specifically, as illustrated in \cref{fig:model}, the input image is augmented with random global transformations to generate two different views $x_1$ and $x_2$. Following BYOL’s example, $x_1$ and $x_2$ are passed through the backbone and projector to generate the projected features $z_1$ and $z_2$. An additional predictor is further utilized to generate $p_1$ to prevent collapse. Since $x_1$ and $x_2$ are from the same image and global transformations do not affect the muscle movements, $p_1$ and $z_2$ should be very similar to each other.

To learn expression-related features in the pre-training stage, we proposed random warping, an unsupervised way to simulate facial muscle movements. We use it to warp $x_1$ to $x_3$, making $x_3$ has a different expression but the same identity and pose as $x_1$. Similar to $z_2$, $z_3$ is extracted and projected from $x_3$ with the same backbone and projector. In order to learn expression-related features, we require $z_3$ and $p_1$ to have a low similarity since they have different expressions. We also find that an additional landmark detection task can help the model to focus on the warped (expression-changed) areas. With these two expression-related pretext tasks, our proposed \modelName{} could empower current self-supervised learning methods to extract expression-related features to benefit downstream FER tasks.

\begin{figure}
  \centering
   \includegraphics[width=0.46\linewidth]{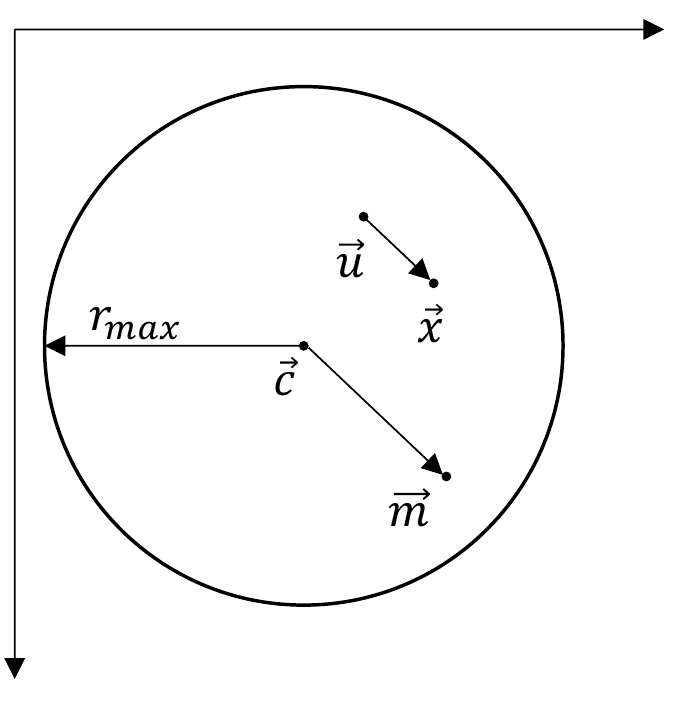}

   \caption{Illustration of the warping process. As point $\vec{c}$ moves to a new position $\vec{m}$, every pixel in the circle should move in the same direction. For example, point $\vec{u}$ moves to $\vec{x}$. That is, the pixel value of the warped face at position $\vec{x}$ can be calculated by finding the before-warping position $\vec{u}$.  }
   \label{fig:warp_diagram}
\end{figure}

\subsection{Face Warping}

To bring expression information to the pre-training phase, we adopt a simple face-warping method~\cite{gustafsson1993interactive} to simulate how facial muscles move when expressing emotions. A facial muscle movement can be seen as the muscle taking a small area of the face around it to move a short distance. We can simulate this process by the above face-warping method. First, we define the warping starting point $\vec{c}$ and ending point $\vec{m}$. To simplify, we assume the warping process only takes effect in a circular area, denoted as a circle with a centre of $\vec{c}$ and a radius of $r_{max}$. All pixels in the circle are supposed to move in the same direction as $\vec{c}$ to $\vec{m}$, but pixels around $\vec{c}$ are supposed to move longer and pixels near the circumference move shorter, making the warping result smooth. For any point $\vec{x}$ in the circle, its content is moved from a source point (denoted as $\vec{u}$). According to~\cite{gustafsson1993interactive}, the source point coordinate vector $\vec{u}$ can be calculated by: 

\begin{equation}
    \vec{u} = \vec{x} - \left(\frac
    {r_{max}^2 - {\lvert\vec{x} - \vec{c} \rvert}^2}
    {r_{max}^2 - {\lvert\vec{x} - \vec{c} \rvert}^2 + {\lvert\vec{m} - \vec{c} \rvert}^2}
    \right)^2\left(\vec{m} - \vec{c}\right)
    \label{eq:warping}
\end{equation}

With \cref{eq:warping}, we can move one ``muscle" efficiently. However, in most cases, emotions are expressed by multiple muscles. To simulate complex expressions, we repeat the above local warping $n$ times with random starting points, radii and moving distances. This allows us to generate various expressions of the same people and backgrounds (as shown in Fig.~\ref{fig:intro}) without any supervision information.

\begin{figure}[t]
  \centering
   \includegraphics[width=0.98\linewidth]{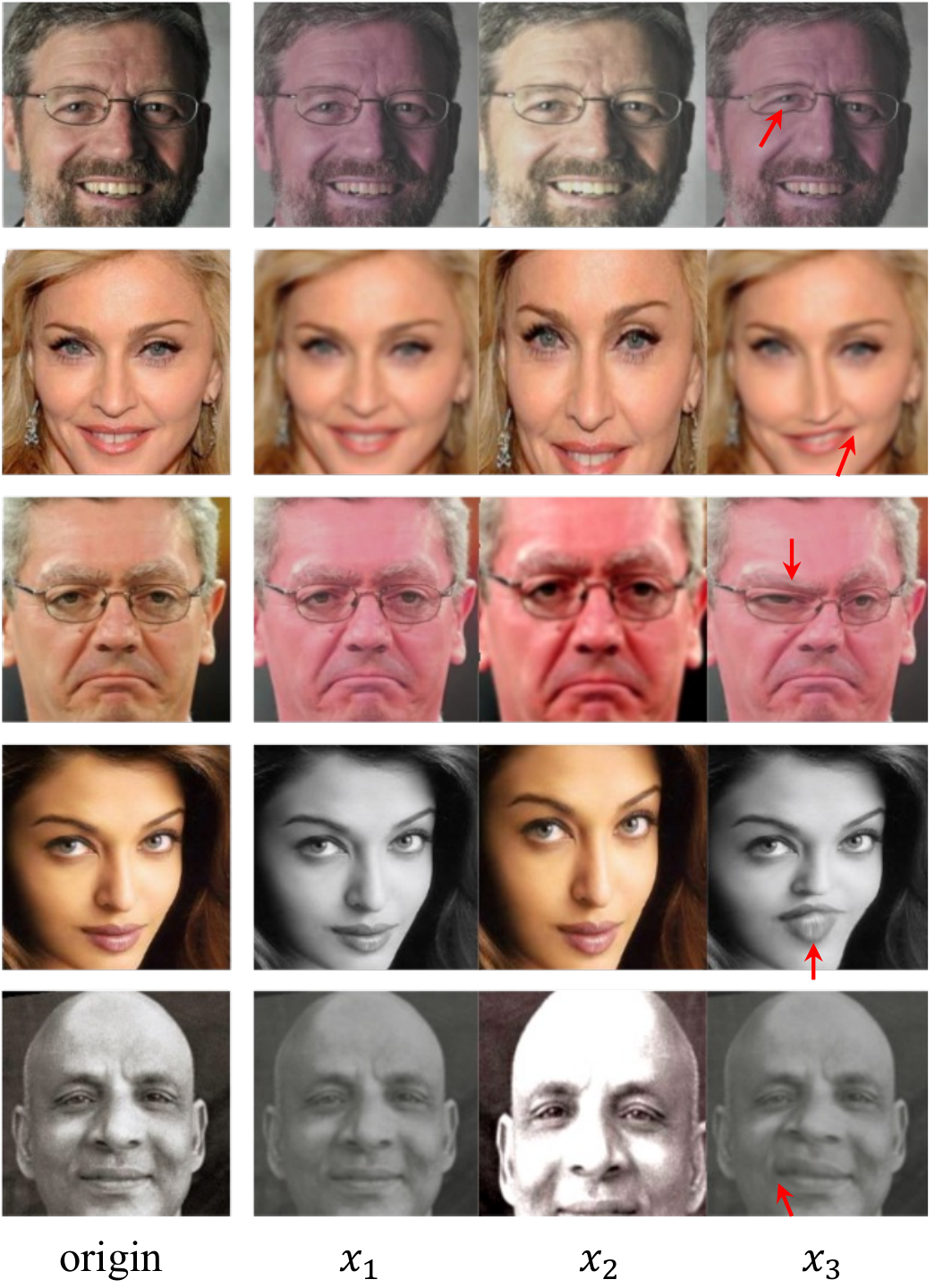}

   \caption{Visualization of the original training images and three branches of input $x_1$, $x_2$ and $x_3$. $x_3$ is randomly warped from $x_1$, and the warped area is marked with a red arrow. (Best zoom in to find details.)}
   \label{fig:vis_ds}
\end{figure}

During training, we generate warped facial images on-the-fly from training samples. We show some examples of warped images from our data loader in \cref{fig:vis_ds}. As shown,  $x_3$ is warped from $x_1$; the warped areas are marked with red arrows for easy identification. Some warped samples are obvious, for example, the slightly opened mouth in the fourth row. Most warping areas are subtle, such as the raised or lowered eyebrow in the first and third rows and the upward mouth in the second and last rows. We assume $x_1$ and $x_3$ have a low similarity. Therefore, obvious warping could help the model converge faster, while subtle warping encourages the model to detect fine-grained muscle movements, which benefit downstream FER tasks.

On the other hand, although our method does not need $x_3$ to change to another emotion, some random warping operation already achieves this. For example, the upward mouth in the second row makes the original \textit{happy} face to \textit{contempt}, and the dropping eyebrow in the third row changes a \textit{sad} face to a slightly \textit{angry} one. These examples show that our warping method could really modify muscle movements and simulate various expressions. We hope this could inspire more interesting methods for FER in the future.

\subsection{Landmark Detection}

To encourage the model to better focus on the warped part of the face, we add a simple landmark detection head with several deconvolutional layers~\cite{zeiler2010deconvolutional} to the framework, as shown in \cref{fig:model}. The landmark detector takes feature maps from $x_1$ and $x_3$ as input and predicts the corresponding landmark points, respectively.

Specifically, the landmark detection model is pre-trained on the 300-W~\cite{sagonas2016300} dataset and is used to directly predict pseudo landmarks for the MS1M dataset. 
The model is lightweight and use HRNetV2-W18~\cite{WangSCJDZLMTWLX19} as its backbone.
It is worth noting that our goal is not to predict accurate landmarks but to help the model find the ``moving muscles" areas as a pretext task. With the predicted pseudo landmark, we perform the same warping with the image $x_1$ to generate the pseudo landmark of $x_3$. The MSE loss is adopted as the criterion. To better push the model to focus on variable parts, we set the loss weight of unchanged landmarks to 0.1 while changed ones to 1.

\subsection{Joint Loss Function}
In our framework, the backbone is jointly trained with the contrastive loss and the landmark detection loss. For contrastive loss, we adopt the symmetrical cosine similarity following SimSiam~\cite{Chen2020SimSIam} and BYOL~\cite{grill2020BYOL}:

\begin{equation}
    sim(i,j) = \frac{1}{2} \left(
    \frac{p_i}{\Vert p_i \Vert} \cdot \frac{z_j}{\Vert z_j \Vert}
    +
    \frac{p_j}{\Vert p_j \Vert} \cdot \frac{z_i}{\Vert z_i \Vert}
    \right)
\end{equation}

Since $x_1$ and $x_2$ are two different views from the same image, so they should have a very high similarity:

\begin{equation}
    L_{cont12} = - sim(1, 2)
\end{equation}

As for $x_3$, it is warped from $x_1$. We hope it is dis-similar from $x_1$ in the expression space. However, since only a small region of the face is changed by warping, we do not want to make them too dissimilar to prevent confusing the model training. Therefore, we set a target similarity (denoted as $s_t$) as a hyper-parameter:

\begin{equation}
    L_{cont13} = max\left(sim(1, 3),\  s_t\right)
    \label{eq:loss_13}
\end{equation}

For landmark detection, the weighted MSE loss is adopted to both $x_1$ and $x_3$:

\begin{equation}
    L_{landmark} = \frac{1}{n} \sum_{i=1}^{n} MSE(w_i \cdot pred_i,\  w_i \cdot pseudo_i)
\end{equation}

where $n$ is the number of landmark points, $w_i$ is the weight for the corresponding point, $pred_i$ and $pseudo_i$ are predicted and the pseudo landmarks, respectively. Then, the joint loss can be formulated as:

\begin{equation}
    L = L_{cont12} + L_{cont13} + \lambda (L_{landmark1} + L_{landmark3})
\end{equation}

where $\lambda$ is a hyper-parameter to balance the losses.

\section{Experiments}
\subsection{Settings}
\textbf{Evaluation tasks.} We evaluate \modelName{} on two facial expression analysis tasks: facial expression recognition and facial expression retrieval. We call them as ``recognition'' and ``retreival'' for brevity. The recognition task is the most popular FEA task and requires predicting discrete categories, while the retrieval task compares expressions in continuous feature space to find the closest one. 

On the recognition task, we follow the standard protocols in SSL and use linear evaluation and k-NN evaluation to measure the quality of extracted features. Specifically, we train one fully-connected layer (or perform k-NN classification) based on features from the \textbf{frozen} backbone. 

On the retrieval task, we evaluate \modelName{} under both direct deployment and fine-tuning scenarios. In the first scenario, we directly use the unsupervised facial expression representation learned from \modelName{} to extract deep features. In the second scenario, the pre-trained models are fine-tuned on the retrieval training set.

\subsection{Datasets}

\textbf{Dataset for training \modelName{}.} We train \modelName{} on 
\textbf{MS1M}~\cite{guo2016ms1m}, a large-scale face recognition dataset with about 3.8M facial images from popular celebrities. Most recognition methods~\cite{vemulapalli2019CompactEmbedding,wang2020suppressing,sheDiveAmbiguityLatent2021,xue2021TransFER,zhang2022LearnAll} use this database for model pre-training. The difference between their pre-training and ours is that they use the supervised face identification task with face ID annotations, while \modelName{} is an SSL method.

\textbf{Dataset for facial expression recognition}. For the recognition task, we use two popular datasets, \emph{i.e.} RAF-DB~\cite{li2017reliable} and {AffectNet}~\cite{mollahosseini2017affectnet}. 
{RAF-DB}~\cite{li2017reliable} is a large-scale FER dataset with 30,000 facial images labelled into seven basic or compound expression categories. Every facial image in this dataset is manually labelled about 40 times to ensure reliability.
{AffectNet}~\cite{mollahosseini2017affectnet} is one of most challenging FER dataset. It consists of about one million facial images collected by searching compression-related keywords on the Internet. Following~\cite{li2021adaptively}, about 280,000 and 3,500 facial images labeled in seven basic categories are adopted for training and testing.

\textbf{Dataset for facial expression retrieval}. For the retrieval task, we use {FEC}~\cite{vemulapalli2019CompactEmbedding}, a large-scale expression comparison dataset by specify the smeariest image pair in each triplet. Since the dataset only release the image url and many urls have been crashed. By removing the broken images, finally about 358K and 28K triplet samples are collected for training and testing, respectively. The triplet prediction accuracy based on extracted features is reported.

\subsection{Implementation Details}
Unless otherwise specified, we utilize a ResNet-18~\cite{he2016deep} as our backbone. For random warping, we random select the starting point $\vec{c}$ from a uniform distribution $U(50, 150)$. Similarly, the moving step ${\lvert\vec{m} - \vec{c} \rvert}^2 \in U(100, 200)$, the radius $r_{max} \in U(50, 80)$, and the strength from $U(150, 300)$. These parameters are for images of size 224 $\times$ 224. The repeat time $n$ is empirically set to 2. For landmark detection, 68 points are inferred from a pre-trained landmark detection model, and the heatmap is generated with a sigma of 1.5.
Other settings for the pre-training and evaluation are described individually below:

\textbf{Pre-training.} As our proposed pretext tasks could combine with various self-supervised learning (SSL) methods. We keep the same settings as the original SSL method by default, except no random crop is adopted for $x_1$ to perform landmark detection. We did not perform tuning on the learning rate or batch size. Specifically, for SimSiam, the SGD optimizer with a 0.05 learning rate is adopted for a mini-batch of 256. The model is pre-trained for 50 epochs on 10\% MS1M for ablation studies and for 20 epochs on 100\% MS1M for comparison with the state of the arts. For BYOL, a big batch size (4096) with the LARS~\cite{you2017large} optimizer is adopted, and the learning rate is set to 4.8. Since the vanilla BYOL need to pre-train for a long while (up to 1000 epochs), we pre-trained it for 50 epochs on 100\% MS1M to compare with SOTA methods.

\subsection{Ablation Studies}

\begin{table}
\centering
\begin{tabular}{lccc}
\toprule

\multirow{2}{*}{Method}   & \multirow{2}{*}{Linear} & \multicolumn{2}{c}{k-NN} \\ \cline{3-4} 
                  &                & 10             &  30           \\ \hline
SimSiam           & 69.95          & 53.10          & 52.54          \\
SimSiam + RW$^*$     & 71.71         & 53.98       &  53.16 \\
SimSiam + RW$^\dag$  & 72.00          & 54.53     &        54.89   \\
SimSiam + RW$^*$ + LD & 73.66          & 55.87   & 56.71  \\  
SimSiam + RW$^\dag$ + LD & \textbf{75.29} & \textbf{62.32} & \textbf{62.97}  \\    
\bottomrule
\end{tabular}
\caption{Evaluation of our proposed pretext tasks with SimSiam on RAF-DB. The top-1 accuracy (\%) is reported. \textbf{RW}: random warping, \textbf{LD}: landmark detection. RW$^*$ represents the totally random warping and RW$^\dag$ represents the landmark-based random warping. See \cref{sec:warping_position} for details. }
\label{tab:ab}
\end{table}

\textbf{Effectiveness of proposed modules.}
In our framework, random warping and landmark detection are two pretext tasks to extract expression-related features for contrastive pre-training methods. To investigate the effect of these two proposed new tasks, we perform an ablation study by pre-training on 10\% MS1M images and evaluating on RAF-DB with three protocols. As the BYOL needs a very large batch size, making it demanding for hardware, we investigated the experiments with SimSiam, which could work with a batch size of 256, which is more resource-friendly.

The results are illustrated in \cref{tab:ab}. As we can see, the vanilla SimSiam could achieve a decent performance: 69.95\% top-1 accuracy by only training a classifier with one FC layer. By applying random warping to the face image, the model can learn from synthetic different expression pairs and outperforms the vanilla SimSiam by a significant margin. Specifically, the proposed random warping increases the linear evaluation performance from 69.95\% to 72.00\$ and boosts the 10-NN and 30-NN performance to 54.53\% and 54.89\%, respectively. We also find that warping based on landmarks performs better than totally random warping, which will be further explored in the following section. With the help of landmark detection, the performance of linear evaluation and 10-NN could further boost to 75.29\% and 62.32\%. These experimental results demonstrate that our proposed random warping could help current contrastive frameworks to learn expression-related features. And the landmark detection task could further help the model to focus on the moving areas. With the help of these two pretext tasks, current SSL models could extract better representations for FER.

\begin{table}
\centering
\begin{tabular}{ccccc}
\toprule
\multirow{2}{*}{$s_t$} & \multirow{2}{*}{$sim(1, 3)$}     & \multirow{2}{*}{Linear} & \multicolumn{2}{c}{k-NN} \\ \cline{4-5} 
        &          &                & 10             &  30           \\ \hline
0    & - 0.16  & 68.58       &   51.50   &   50.85      \\
0.2  & 0.08   & 69.95       &   50.55   &   50.29       \\
0.4  & 0.26   & 73.99       &   55.34   &   56.45     \\ 
0.6  & 0.45   & \textbf{75.00} & \textbf{55.87}   & \textbf{56.71}     \\
0.8  & 0.69   & 73.21       &   55.08   &   55.67       \\
\bottomrule
\end{tabular}
\caption{Exploring the target similarity $s_t$ between $x_1$ and $x_3$ on RAF-DB with SimSiam. For generality, the warping position is \textbf{randomly selected} in these experiments. }
\label{tab:target_similar}
\end{table}

\textbf{Target similarity $s_t$ between $x_1$ and $x_3$.}
Our proposed \modelName{} utilizes the random local warping to simulate facial muscle movements. The basic hypothesis is that the warped face ($x_3$ in \cref{fig:model}) has a different expression from the original face ($x_1$ in \cref{fig:model}). To distinguish different expressions, we push $x_1$ and $x_3$ to have a low similarity. However, except for the warped area, facial muscles in other regions share the same status with the original $x_1$. To prevent confusing the learning process, we set a target similarity $s_t$ as a lower bound as described in \cref{eq:loss_13}. The lower $x_t$ will push the model to focus more on moving areas and ignore other regions, which is harmful to the model to pull $x_1$ and $x_2$ close. It's a trade-off between these two tasks.

Results with different $s_t$ values are illustrated in \cref{tab:target_similar}. Surprisingly, we find that although we have set the lower bound of $sim(1,3)$ by $s_t$, the observed $sim(1,3)$ is still a little lower than $s_t$. This may be because the punishment still exists when at least one sample has a higher similarity in the mini-batch, making the overall mean similarity lower. To better illustrate the learning result, we also report the mean $sim(1,3)$ of the last mini-batch at the pre-training stage in \cref{tab:target_similar}. As we can see, the framework achieves the best performance (75\% for linear evaluation) when $s_t$ is set to 0.6 and the actual $sim(1,3)$ is 0.45. When $s_t$ is set to 0, the model tries to represent various permutations of muscle movements as orthometric, which ignores the common areas and performs worst (68.58\%) for linear evaluation and is relatively poor for k-NN classification. When $s_t$ is set to a high value, \eg 0.8, the model has less incentive to distinguish different muscle movements and can not extract effective expression presentations.

\textbf{Warping Position: Random V.S. Landmark-based.}
\label{sec:warping_position}
\begin{table}
\centering
\begin{tabular}{cccc}
\toprule
\multirow{2}{*}{Method} & \multirow{2}{*}{Linear} & \multicolumn{2}{c}{k-NN} \\ \cline{3-4} 
                        &                & 10             &  30           \\ \hline
Random           &  75.00           &   55.87   &   56.71   \\
Landmark-based   & \textbf{75.29}   & \textbf{62.32} &  \textbf{62.97}   \\
\bottomrule
\end{tabular}
\caption{Comparison of random landmark-based warping on RAF-DB with SimSiam.}
\label{tab:warping_position}
\end{table}

As random warping is proposed to simulate facial muscle movements and change the expression, we hope it could take effect on the physiological facial muscle areas. However, warping on irrelevant areas, such as the forehead, cheek, or hair areas, may not affect the expression and may introduce false negative samples, which is harmful to the learning process. An intuitive way is to only perform warping around facial landmarks. 
\cref{tab:warping_position} shows the performance comparison of totally random and landmark-based warping methods. The landmark-based method performs better in all three protocols. Moreover, the landmark-based random warping is more helpful in boosting k-NN performance: increasing from 55.87\% and 56.71\% to 62.32\%, and 62.97\% for k = 10 and 30, respectively, while marginal improvement (0.29\%) for linear evaluation.

Although the landmark-based method performs better, the total random warping procedure performs comparatively and is more flexible in applying to other no-landmark-available scenes. We hope both methods can inspire researchers to design more interesting works.

\subsection{Combining with various SSL methods.}

Our random warping could combine with various contrastive SSL methods.
In principle, our proposed modules could help existing SSL methods to focus on the muscles that move during facial expression, and better SSL methods could extract more robust features.

We conduct experiments with three SOTA SSL methods to investigate the compatibility with our proposed modules in \cref{tab:ssl_method}.
We also compare them with random initialization and supervised training on the MS1M dataset. As shown, SimSiam and BYOL outperform the supervised ones without any annotation. This is because the face recognition pre-train restrains the model from learning expression features. By introducing expression-related tasks in the pre-training stage, we improve the performance of all three SSL methods. The improvement with MoCo V2 is less significant because it does not have a $s_t$ hyper-parameter to balance the similarity between $x_1$ and $x_3$. We simply append our $x_3$ feature to the dictionary as full negative examples. However, we still increase the linear evaluation accuracy of MoCo V2 by about 1\%. Moreover, these results suggest that our method can benefit from a stronger SSL method to perform better.

\textbf{Comparison with general facial representation learning methods.} Some methods~\cite{li2019TCAE,chang2021LearningFacial,bulat2022Pretrainingstrategies} aim to extract universal facial representations with identity, pose, expression, and even landmark information while our method only focuses on expression features. As shown in Tab.~\ref{tab:general}, our method outperforms these general methods by a big margin: under linear evaluation, our method outperforms FaceCycle~\cite{chang2021LearningFacial} with 8.13\%  with a Res-18 backbone, and outperforms Flickr-Face~\cite{bulat2022Pretrainingstrategies} with more than 4\% with Res-50, indicating our overwhelming superiority over general methods. 

\begin{table}
\centering
\begin{tabular}{lccc}
\toprule
\multirow{2}{*}{Method}   & \multirow{2}{*}{Linear} & \multicolumn{2}{c}{k-NN} \\ \cline{3-4} 
                  &                & 10             &  30           \\ \hline
Random Init                     & 51.43 & 44.49 &   45.21 \\
Sup. (MS1M)                     & 67.47 & 52.15 &   53.32   \\ \hline
MoCo V2~\cite{chenMoCoV22020}   & 57.82 & 44.10 &    \textbf{46.48}    \\
\rowcolor{pearDark!20} MoCo V2 + Ours  & \textbf{58.77} & \textbf{44.91} & 45.66  \\ \hline
SimSiam~\cite{chen2020exploring}& 69.33 & 60.00 &   60.30   \\
\rowcolor{pearDark!20}SimSiam + Ours &  \textbf{75.32} &  \textbf{62.57} &  \textbf{63.39}   \\ \hline   
BYOL~\cite{grill2020BYOL}       & 77.09 & 61.04 &   62.06   \\
\rowcolor{pearDark!20}BYOL + Ours    &  \textbf{79.95} &  \textbf{62.48} &    \textbf{63.92}   \\
\bottomrule
\end{tabular}
\caption{Comparison of our framework with other state-of-the-art self-supervised learning methods on RAF-DB.}
\label{tab:ssl_method}
\end{table}

\begin{table}
\centering
\begin{tabular}{ccccc}
\toprule
\multirow{2}{*}{Method} & \multirow{2}{*}{Backbone} & \multirow{2}{*}{Linear} & \multicolumn{2}{c}{k-NN} \\ \cline{4-5} 
                        &     &           & 10             &  30           \\ \hline
TCAE~\cite{li2019TCAE}          &  9-layer  &  65.32   &  59.19   & 57.98  \\
FaceCycle~\cite{chang2021LearningFacial} & 16-layer & 71.01  & 55.80 &  55.80 \\
\rowcolor{pearDark!20}Ours   &  Res-18  &  \textbf{79.95} & \textbf{62.48}  &  \textbf{63.92} \\ \hline
Flickr-Face~\cite{bulat2022Pretrainingstrategies} & Res-50  & 80.70 & 60.01  & 60.36 \\
\rowcolor{pearDark!20}Ours   &  Res-50  &  \textbf{84.64} & \textbf{66.72}  &  \textbf{68.06} \\
\bottomrule
\end{tabular}
\caption{Comparison of our framework with other state-of-the-art general facial representation learning methods on RAF-DB.}
\label{tab:general}
\end{table}

\textbf{Comparison with state-of-the-art FER methods.}
We also find that the baseline strategy, which uses a pure Res18 network with only random crop and flips as data augmentation, achieves a comparable performance for FER with our \modelName{}. As illustrated in \cref{tab:sota}, the baseline performs similarly (89.18\%) with RUL (88.98\%) on RAF-DB and ranks second (64.94\%) on AffectNet, demonstrating the great potential of our method to boost FER performance.

\begin{table}
\centering
\begin{tabular}{@{}cccc@{}}
\toprule
Method          & Year & RAF-DB & AffectNet \\ \midrule
IPA2LT~\cite{zeng2018facial}          & 2018 & 86.77  & 57.31     \\
RAN~\cite{wang2020region}             & 2020 & 86.90  & 59.50     \\
SCN~\cite{wang2020suppressing}        & 2020 & 87.03  & 60.23     \\
KTN~\cite{li2021adaptively}           & 2021 & 88.07  & 63.97     \\
DMUE~\cite{sheDiveAmbiguityLatent2021}& 2021 & 88.76  & 62.84     \\
RUL~\cite{zhang2021RelativeUncertaintya}& 2021 & 88.98  & 61.43     \\
Face2Exp~\cite{zeng2022Face2Exp}     & 2022 & 88.54  & 64.23     \\
EAC~\cite{zhang2022LearnAll}             & 2022 & \textbf{89.99}  & \textbf{65.32}     \\ \midrule
\rowcolor{pearDark!20} Baseline + Ours & 2023 & 89.18  & 64.94  \\
\bottomrule
\end{tabular}
\caption{Comparison with our baseline strategy with current state-of-the-art FER methods.}
\label{tab:sota}
\end{table}

\subsection{Image Retrieval.}
Image retrieval is another application that could benefit from continuous expression representation. For example, belly laughs and smiles may take place in different scenes, although they have the same basic category: happy. To investigate the effect of our proposed \modelName{}, we reported the performance on the FEC database without and with fine-tuning. As shown in \cref{tab:image_retrieval}, our \modelName{} outperforms MS1M supervised weights significantly, from 34.78\% to 39.78\%. Even after fine-tuning, our method still performs better, reaching up to 78.31\%. This indicates that our unsupervised approach can learn continuous expression features and benefit image retrieval.

\begin{table}
\centering
\begin{tabular}{clcc}
\toprule
Fine-tuning          & Embedding      & $l2$      & cos  \\ \midrule
\multirow{4}{*}{No}  & Random Init    & 34.40   & 33.75 \\
                     & MS1M           & 34.78   & 33.56 \\
                     & \cellcolor{pearDark!20}  Ours    & \cellcolor{pearDark!20} \textbf{39.78}   & \cellcolor{pearDark!20} \textbf{43.59} \\ 
\midrule
\multirow{4}{*}{Yes} & Random Init    & 38.37   & 38.38  \\
                     & MS1M           & 77.52   & 76.99  \\
                    & \cellcolor{pearDark!20} Ours      & \cellcolor{pearDark!20} \textbf{78.31}   & \cellcolor{pearDark!20} \textbf{77.30} \\
\bottomrule
\end{tabular}
\caption{Comparison of different pre-trained weights on FEC.}
\label{tab:image_retrieval}
\end{table}

\section{Conclusion}

In this paper, we propose a novel method (\modelName{}) for unsupervised facial expression representation (UFER) learning.
The key point of \modelName{} is leveraging local warping to generate expressive variations of face images.
We use contrastive learning and landmark detection as two pretext tasks to learn from the locally-warped images.
In-depth investigations on expression recognition and retrieval tasks show that the \modelName{} representation has gained a considerable discriminative ability for facial expression analysis. 
Moreover, we demonstrate that \modelName{} can be used as an effective pre-training strategy that outperforms popular pre-training with the face identification pretext task.

{\small
\bibliographystyle{ieee_fullname}
\bibliography{egbib}
}

\end{document}